\documentclass[10pt,twocolumn,letterpaper]{article}

\usepackage[pagenumbers]{iccv}      
\usepackage{subcaption}
\usepackage{pgfplots}
\pgfplotsset{compat=1.17}
\usepackage{booktabs} 
\usepackage{multirow}
\usepackage[ruled]{algorithm2e} 
\usepackage[dvipsnames]{xcolor}
\newcommand{\yes}{\textcolor{ForestGreen}{\checkmark}}
\newcommand{\no}{\textcolor{red}{\ding{55}}}
\usepackage{pifont}

\SetAlFnt{\small}
\SetAlCapFnt{\small}
\SetAlCapNameFnt{\small}
\SetAlCapHSkip{0pt}
\SetAlgoNoEnd

\usepackage{graphicx} 

\definecolor{iccvblue}{rgb}{0.21,0.49,0.74}
\usepackage[pagebackref,breaklinks,colorlinks,allcolors=iccvblue]{hyperref}

\def\paperID{12952} 
\def\confName{ICCV}
\def\confYear{2025}

\title{Cabbage: A Differential Growth Framework for Open Surfaces}

\author{
\begin{tabular}{ccc}
Xiaoyi Liu$^{1, *}$ \quad Hao Tang$^{2,\dagger}$\\
$^{1}$Washington University in St. Louis \quad $^{2}$Peking University \\
{\tt\small jasonl@wustl.edu} \quad {\tt\small haotang@pku.edu.cn}
\end{tabular}
}

\begin{document}
\twocolumn[{
\maketitle
\begin{center}
    \captionsetup{type=figure}

    \begin{minipage}{\textwidth}
        \centering
        \begin{minipage}{0.24\textwidth}
            \includegraphics[width=\linewidth]{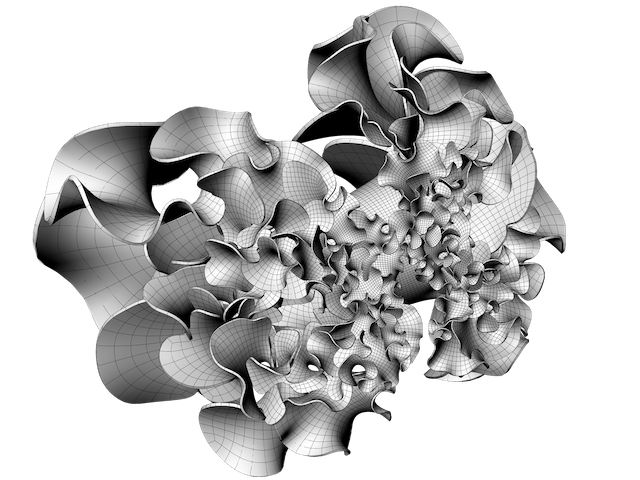}
        \end{minipage}
        \hfill
        \begin{minipage}{0.24\textwidth}
            \includegraphics[width=\linewidth]{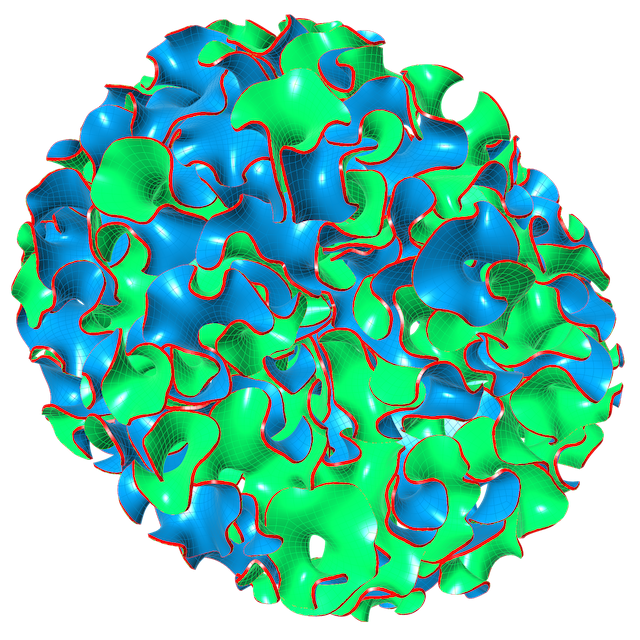}
        \end{minipage}
        \hfill
        \begin{minipage}{0.24\textwidth}
            \includegraphics[width=\linewidth]{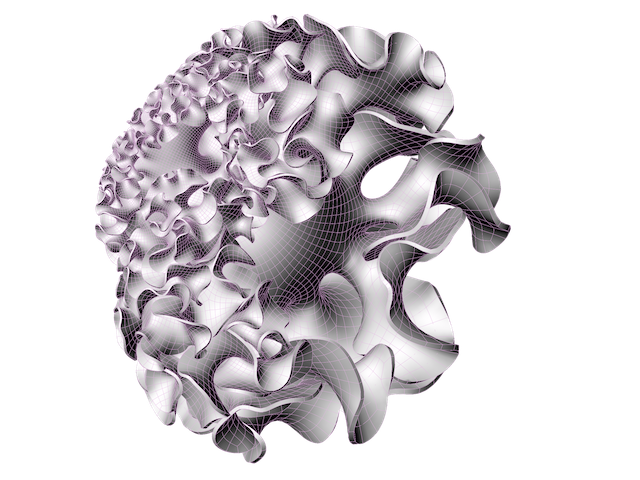}
        \end{minipage}
        \hfill
        \begin{minipage}{0.24\textwidth}
            \includegraphics[width=\linewidth]{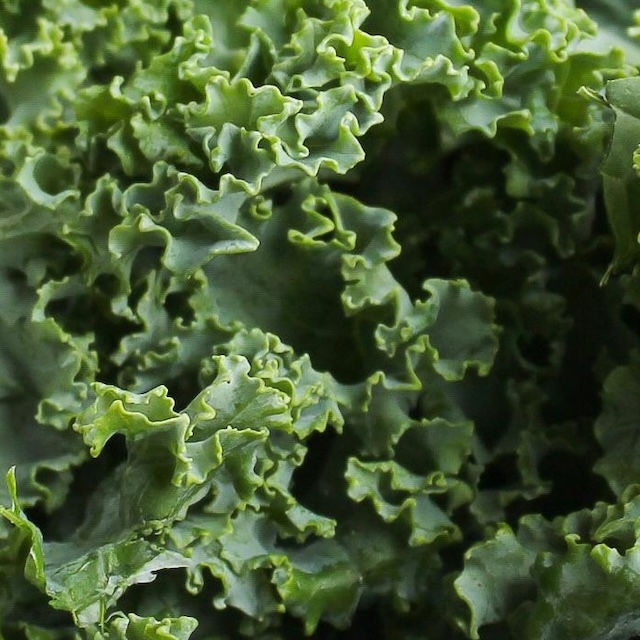}
        \end{minipage}
        
        \vspace{0.3cm} 
        \begin{minipage}{0.24\textwidth}
            \includegraphics[width=\linewidth]{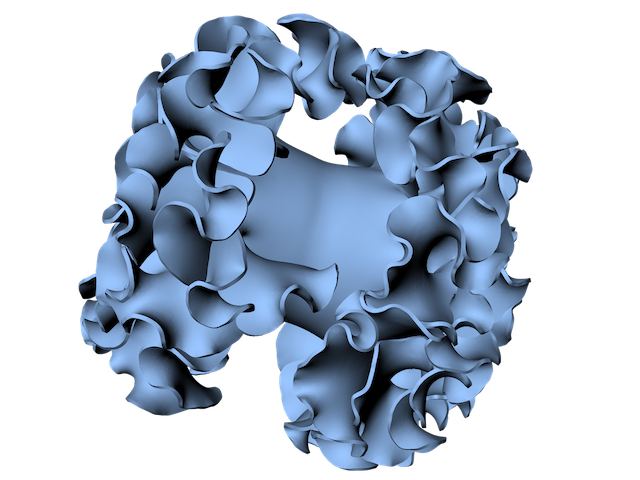}
        \end{minipage}
        \hfill
        \begin{minipage}{0.24\textwidth}
            \includegraphics[width=\linewidth]{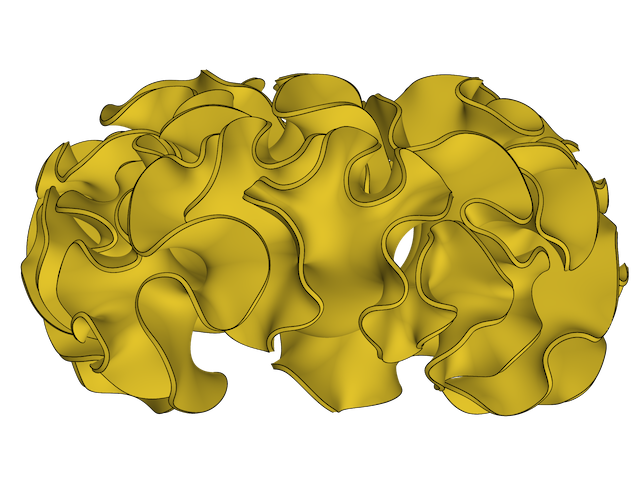}
        \end{minipage}
        \hfill
        \begin{minipage}{0.24\textwidth}
            \includegraphics[width=\linewidth]{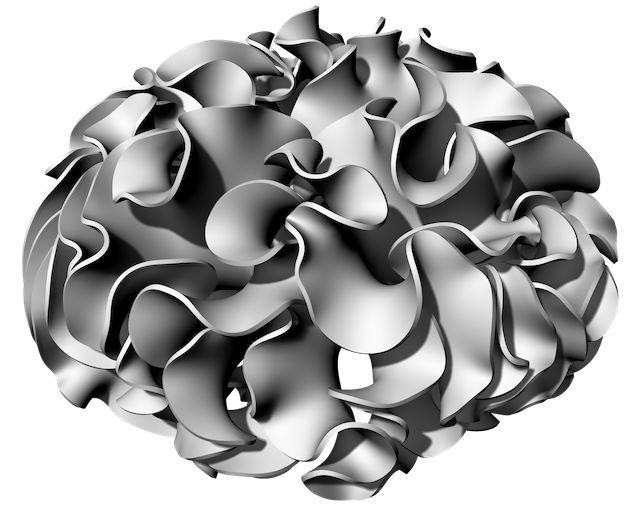}
        \end{minipage}
        \hfill
        \begin{minipage}{0.24\textwidth}
            \includegraphics[width=\linewidth]{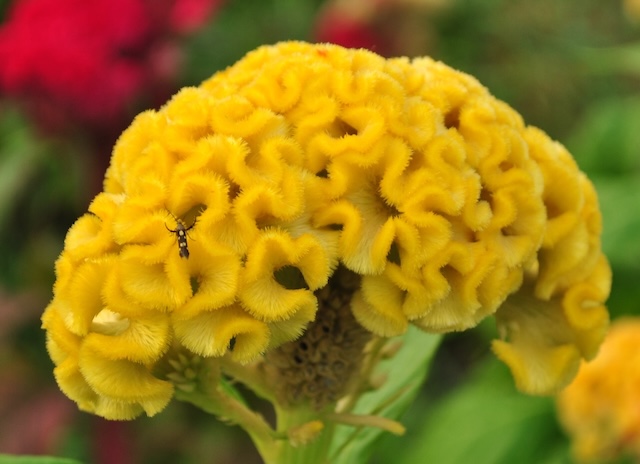}
        \end{minipage}
    \end{minipage}

    \caption{(Left $3$ columns) Generated CAD models vs. photos of kale and Celosia flower (right-most column). Each column shows a different variation of structure and curvature.}
    \label{fig:teaser}

\end{center}
}]


\renewcommand{\thefootnote}{} 
\footnotetext{$^\dagger$Corresponding author.\\ \indent \indent $^*$Work Done during the visit at Peking University.}
\renewcommand{\thefootnote}{\arabic{footnote}} 

\begin{abstract}
We propose Cabbage, a differential growth framework to model buckling behavior in 3D open surfaces found in nature-like the curling of flower petals. Cabbage creates high-quality triangular meshes free of self-intersection. Cabbage-Shell is driven by edge subdivision which differentially increases discretization resolution. Shell forces expands the surface, generating buckling over time. Feature-aware smoothing and remeshing ensures mesh quality. Corrective collision effectively prevents self-collision even in tight spaces. We additionally provide Cabbage-Collision, and approximate alternative, followed by CAD-ready surface generation. Cabbage is the first open-source effort with this calibre and robustness, outperforming SOTA methods in its morphological expressiveness, mesh quality, and stably generates large, complex patterns over hundreds of simulation steps. It is a source not only of computational modeling, digital fabrication, education, but also high-quality, annotated data for geometry processing and shape analysis. \end{abstract}
\begin{figure*}[t]  
  \centering
 \includegraphics[width=\textwidth,keepaspectratio]{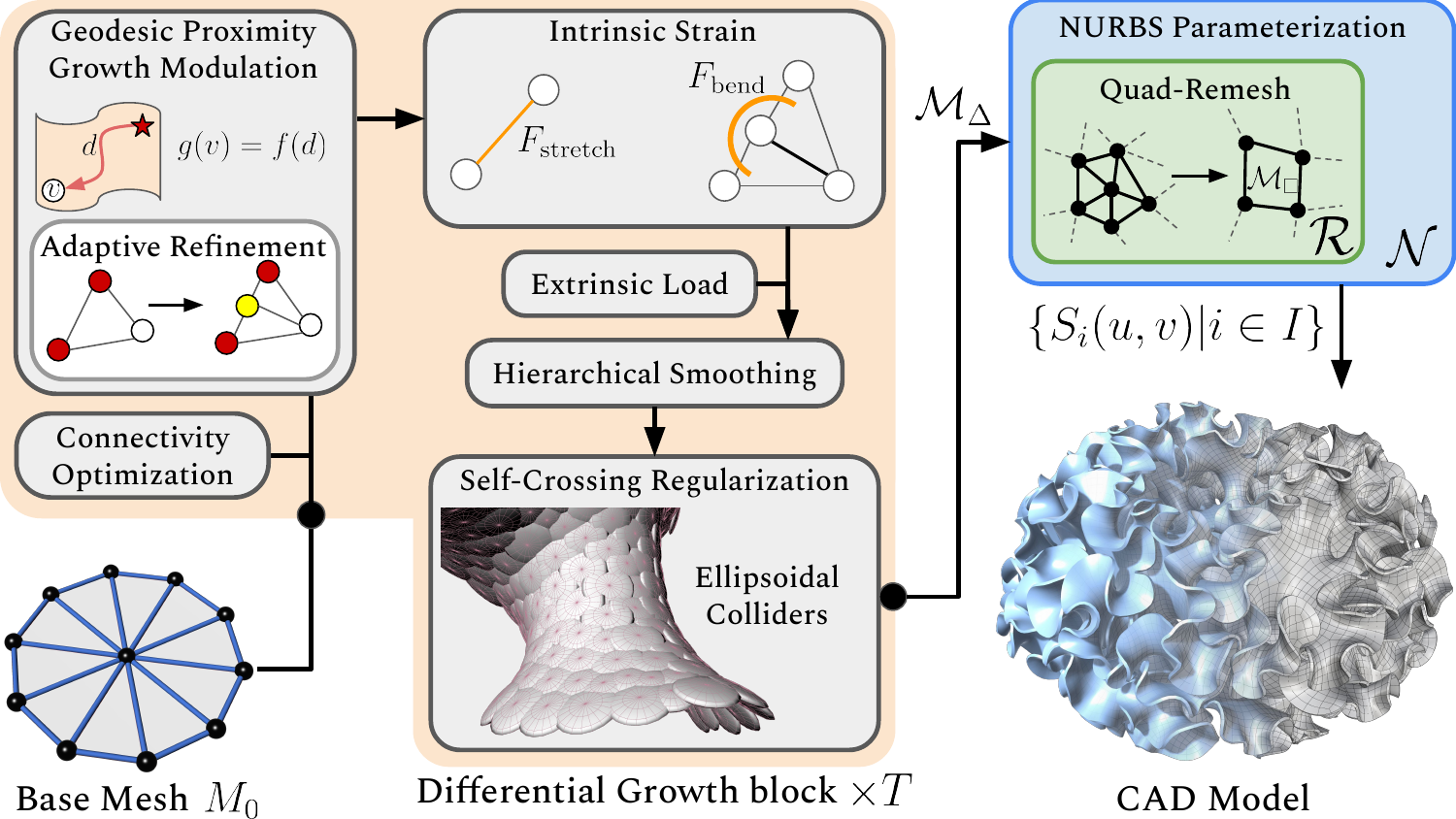}  
    \caption{The proposed Cabbage pipeline is composed of $T$ iterative applications of the Differential Growth block to an initial mesh $M_0 = (\mathcal{V}, \mathcal{E}, \mathcal{F})$ followed by a CAD model conversion step. In the former, mesh connectivity is optimized with edge flips, collapses, and ear removal. Growth factors $g(v) \forall v\in\mathcal{V}$ are computed as a function Eq~\ref{eq:growth-factor} of geodesic distance to select vertices of the mesh, and they drive differential edge subdivision. The mesh surface is deformed by bending and stretching forces and optional external fields. Global and local feature-aware smoothing ensures mesh fairness. Vertex-vertex collisions prevent self-intersections. After desired morphology develops, the triangular mesh $M_\Delta$ is quadrangulated, offset, and parametrized by a polysurface composed of individual NURBS patches $\{S_i(u, v)|i \in I \}$}
    \label{fig:arch}
    \vspace{-0.4cm}
\end{figure*}

\section{Introduction}
\label{sec:intro} Curled organic surfaces in Nature, such as flower petals, leaves, and fungus, inspired researchers and artists alike. These emergent shapes has been credited to differential growth, the process where growth occur at varying rates across a medium. Computationally reproducing this complex process can greatly enrich the generation and fabrication of artistic products. Existing approaches leave room for improvement in robustness, simplicity, and capacity for interactive, art-directed adaptations.

Open surface representations with triangular meshes is a natural choice to model thin, flexible structures like flower petals. 2D surfaces embedded in 3D space provides structure for growth and buckling while avoiding unnecessary computational complexity of explicitly representing the thickness of these very thin surfaces. Preventing self-intersections is critical for up-close visualization, geometric operations, and manufacturability. 

However, Computer-Aided Design (CAD) workflows often require parametric representations such as NURBS (Non-Uniform Rational B-Splines) \cite{NURBS} to precisely carry out editing operations like boolean. Offsetting high-curvature NURBS surfaces to create thickness often results in self-intersections or invalid geometry.

Our method, Cabbage, addresses these challenges by providing a novel pipeline that integrates differential growth with user-directed controls and CAD-ready output. Our key contributions are:
\begin{itemize}
    \item Simple and Robust Differential Growth Framework: Cabbage models curling with shell forces, maintaining a smooth, nearly isotropic and 5-6-7 mesh free of self-intersection as complex patterns develop over hundreds of simulation steps.
    \item Art-Direction and Morphological Range: Intuitive parameters such as bending stiffness and the distribution of growth factors weave a wide shape space. From petals that bend gently over large areas or curl tightly and repeatedly in small spaces, Cabbage enriches creative exploration without the full computational burden of physics-based solvers that rely on complex force derivatives.
    \item Seamless CAD Integration: The pipeline converts the open surface mesh into a smooth, manufacturable NURBS representation using quad retopology. This ensures compatibility with industry workflows, producing valid NURBS surfaces even in high-curvature regions.

\end{itemize}
By combining differential growth dynamics with intuitive user control and CAD integration, Cabbage bridges the gap between organic aesthetics and engineering precision, elegantly balancing realism and speed to offer a powerful tool for computational design and digital fabrication.

\begin{table}[t]
\centering
\resizebox{\columnwidth}{!}{
\begin{tabular}{lcccccccccc}
\toprule
 & (1) & (2) & (3) & (4) & (5) & (6) & (7) & (8) & (9) & (10)\\
\midrule
Bachman \cite{Bachman} & \yes & \no & \no & \no & \no & \no & \no & \no & \no & \no\\
Horikawa \cite{Horikawa} & \yes & \yes & \no & \no & \no & \no & \no & \no & \no & \yes\\
Okunskiy \cite{Okunskiy} & \yes & \no & \yes & \no & \no & \no & \no & \no & \yes & \no\\
Yu \cite{EmYu} & \yes & \no & \no & \yes & \no & \no & \no & \no & \yes & \yes\\
FloraForm \cite{FloraForm} & \no & \yes & \yes & \yes & \no & \no & \no & \no & \no & \yes\\
\textbf{Cabbage (ours)} & \yes & \yes & \yes & \yes & \yes & \yes & \yes & \yes & \yes & \yes\\
\bottomrule
\end{tabular}
}
\caption{Comparison of features: (1) Has open-sourced implementation. (2) Multiscale Feature Generation. (3) Does not rely on proprietary/paid software. (4) Has mechanism to prevent self-intersection. (5) Has CAD software integration. (6) Programmable growth dynamics. (7) Documentation and tutorial for users without a technical background. (8) Logging and Error-catching. (9) Lightweight code. (10) Mesh quality and fairness optimization.}
\vspace{-0.4cm}
\end{table}

\section{Related Work}
\textbf{Mechanistic and Procedural Modeling of Differential Growth}. The development of complex morphology over time in nature, from the buckling of leaves to the curling of plant organs has long inspired research \cite{Lindenmayer:1968}. Such self-organization has been attributed to local interactions like as stress \cite{Trinh-MechanicalForces, Lebovka-RadialGrowth}, strain \cite{Huang-PlantOrgans, Walia-CurvedPlantOrgans}, and hinge forces \cite{Guo-DehydrationFolding} by plant cells on the material that leads to global behavior like bending, stretching, and folding \cite{Liang:2011BloomingLily}. Differential impact of growth-inducing substances such as auxins, measured as growth rates \cite{Zadnikova-ApicalHookFormation, Hong-AncientPlantForm} further diversify morphology. Unfortunately, code availability and need for domain expertise limit the application beyond biological settings.

In visual computing, mechanistic interactions has been modeled by shell forces \cite{Grinspun:2003DiscreteShells} and harnessed in programmable materials \cite{Lukovic-Auxetics}, robotics \cite{Yang-MorphingMatter}, and mimicry of the natural forms that inspired them \cite{Owens-inflorescence}. Modeling and generating natural forms for aesthetic, design, and digital fabrication purposes remains an active area of research. Cabbage build on the theoretical foundation of these efforts by providing a robust framework for generating 3D open surfaces with a wide range of curled morphology. Our lightweight Python implementation of Cabbage balances technical grounding and accessibility to a wider audience, from computer scientist outside the computer graphics domain to artists, providing a resource of generating diverse and high-quality mesh surfaces for computational design, art-making, computer vision, and learning.

\noindent \textbf{Applied Differential Growth}. Differential growth has been exploited to generate artistic forms by leveraging advances in shell physics and 3D modeling.

Nervous System studio \cite{FloraForm} produced compelling visual works and fabricated pieces by simulating growth with stretch and bending forces as described in \cite{Grinspun:2003DiscreteShells}. Their modeling of growth rates as a function of the geodesic distance to a user-defined 'source' is a simple and effective model for differential material property. In an overdamped implicit Euler solver, employs adaptive edge-length subdivision for complex buckling, and maintains mesh quality through edge flips, while ellipsoid colliders on tangent planes prevent self-intersections. Nonetheless, the intricate force computations and the project’s closed-source nature restrict community engagement.

Open-source methods attempted further simplification. Hoff \cite{Inconvergent} modeled vertex movement using attraction from neighboring vertices and repulsion from those within a specified distance, with binary growth rates smoothed over local neighborhoods. Okunskiy’s plugin \cite{Okunskiy} adopts a similar scheme through Blender's \cite{Blender} interface. However, severe anistropism and sharp curvature limits applications beyond rendering. Yu’s web app \cite{EmYu} advances the collision-based approach introducing external forces like gravity and maintaining mesh quality barycentric smoothing. However, self-intersection persists in tight spaces, and the range of morphology is limited.

Alternative approaches were explored by Horikawa \cite{Horikawa} in Houdini \cite{Houdini} by perturbing the mesh and increasing resolution through global isotropic remeshing. However, buckling occur only at large scale and the mesh remains crumpled due to the coarse base mesh. Bachman \cite{Bachman} explored dynamic relaxation \cite{DynamicRelaxation} in Rhino \cite{Rhino, Kangaroo} with sphere colliders, but could not maintain smoothness as the mesh growth large without manual adjustments to collider radius during the simulation. The proprietary nature of these software further limits access.

Cabbage extends these collision-based paradigms by decoupling the simulation of surface dynamics from self-intersection prevention. In addition to smoothing, tunable bending forces model the surface’s resistance to deformation. Unlike all previous methods, our pipeline additionally converts the grown mesh into a closed, CAD-ready NURBS volume with creased edges. The open-source Python implementation provides designers, artists, and educators with a robust and accessible resource for exploring curled morphologies and learning how to model them.

\begin{figure*}[ht!]
    \centering
    \setlength{\tabcolsep}{2pt}
    \renewcommand{\arraystretch}{0.5}
    \begin{tabular}{cccccc}
        \includegraphics[width=0.15\textwidth]{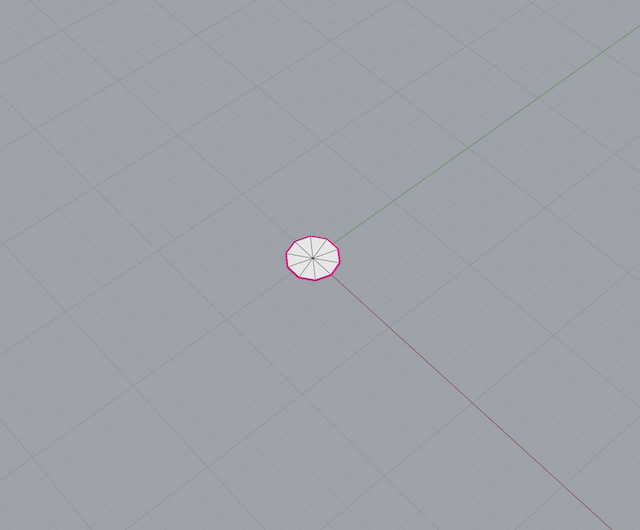} &
        \includegraphics[width=0.15\textwidth]{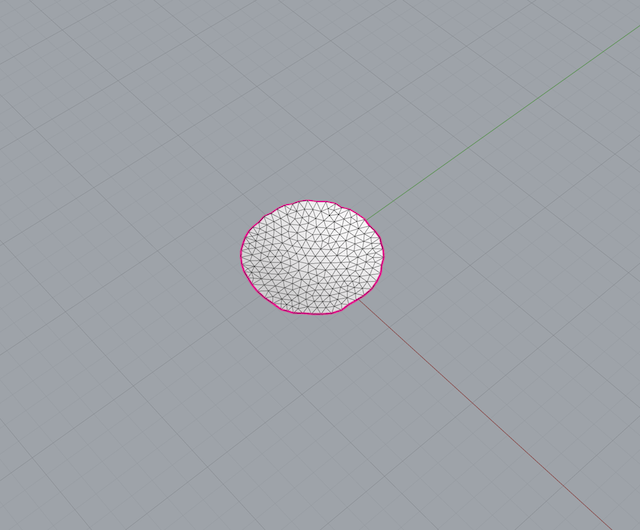} &
        \includegraphics[width=0.15\textwidth]{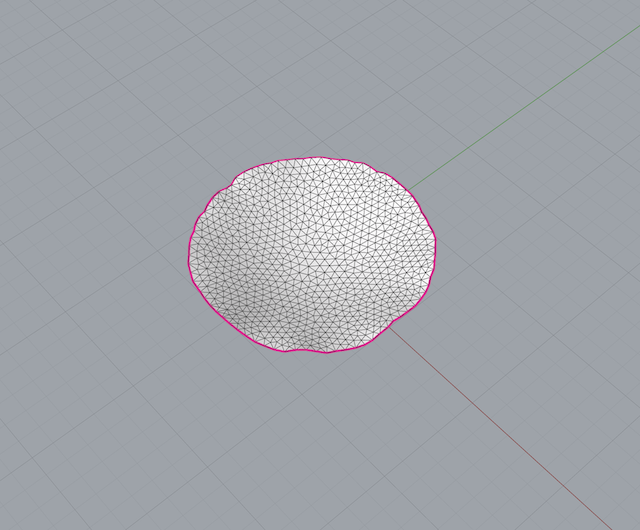} &
        \includegraphics[width=0.15\textwidth]{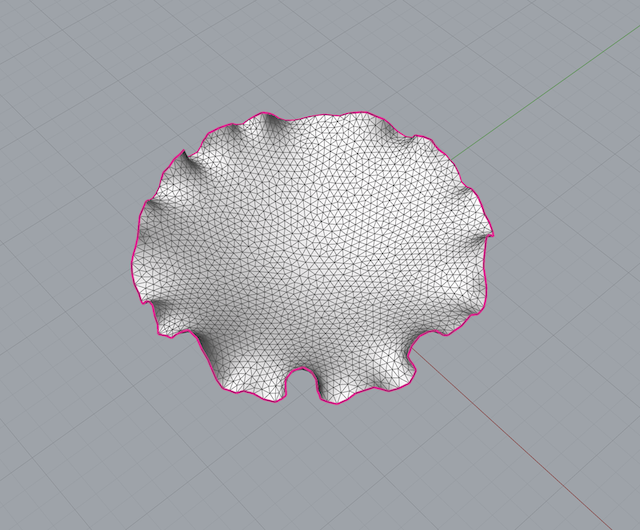} &
        \includegraphics[width=0.15\textwidth]{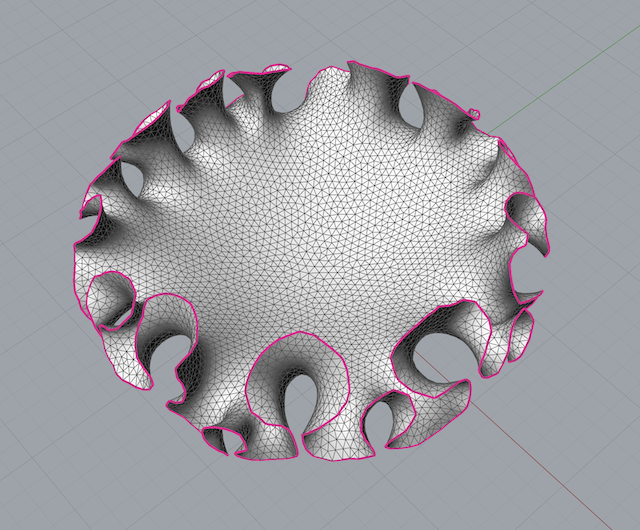} &
        \includegraphics[width=0.15\textwidth]{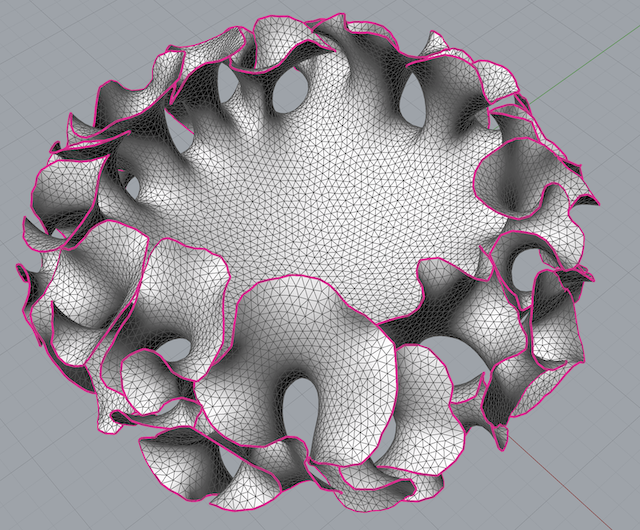} \\
        \includegraphics[width=0.15\textwidth]{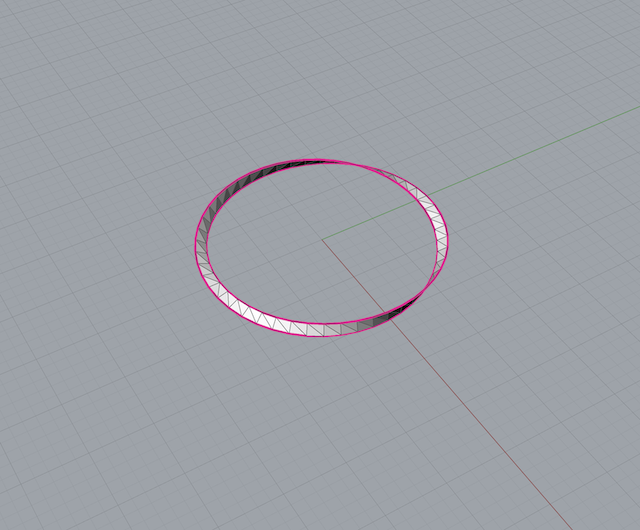} &
        \includegraphics[width=0.15\textwidth]{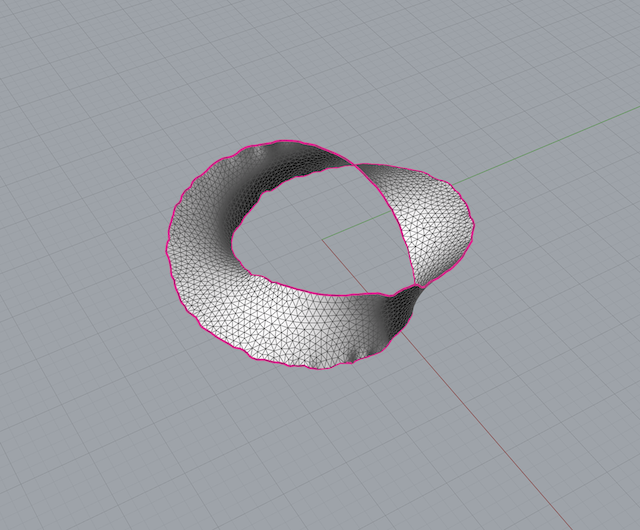} &
        \includegraphics[width=0.15\textwidth]{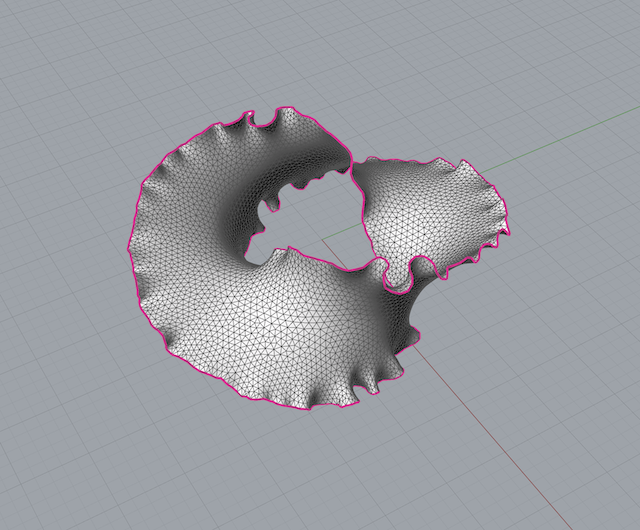} &
        \includegraphics[width=0.15\textwidth]{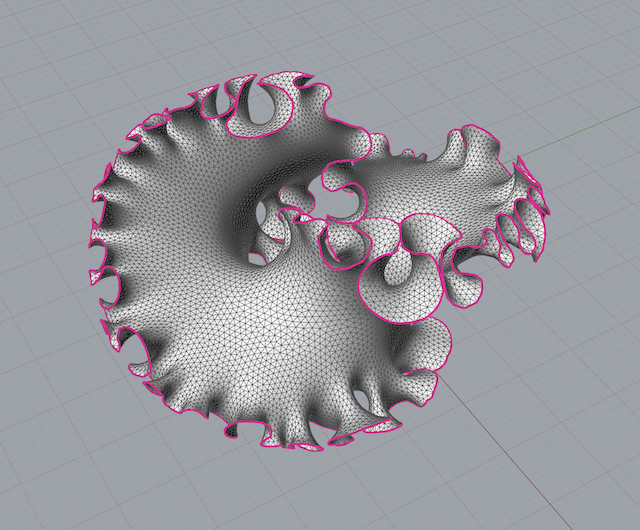} &
        \includegraphics[width=0.15\textwidth]{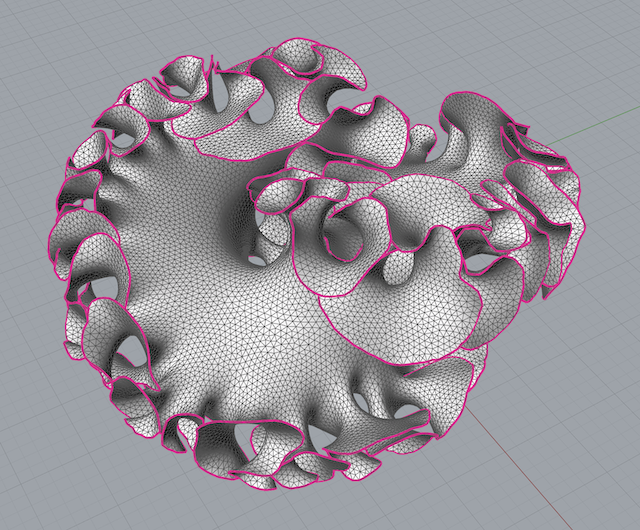} &
        \includegraphics[width=0.15\textwidth]{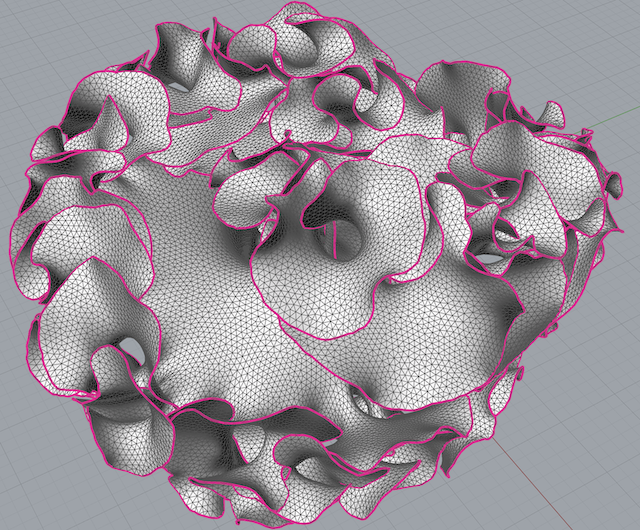} \\
        \includegraphics[width=0.15\textwidth]{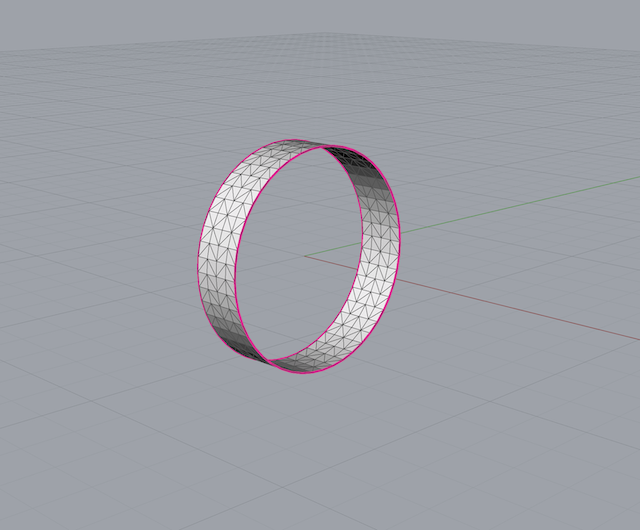} &
        \includegraphics[width=0.15\textwidth]{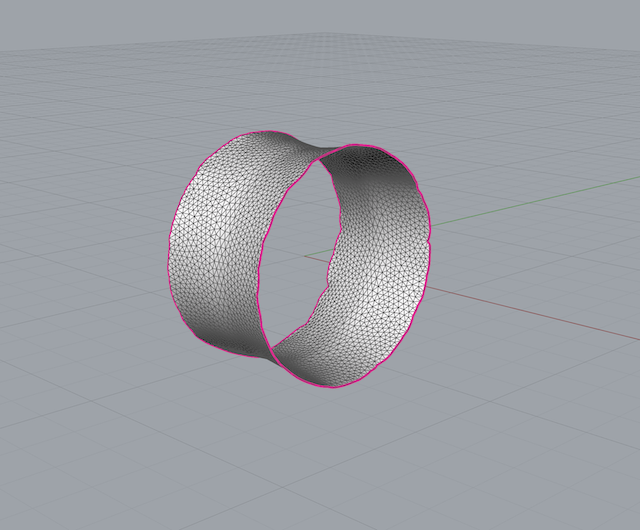} &
        \includegraphics[width=0.15\textwidth]{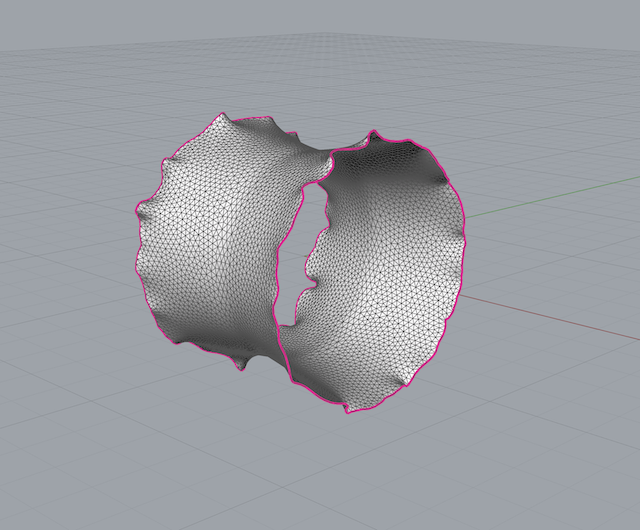} &
        \includegraphics[width=0.15\textwidth]{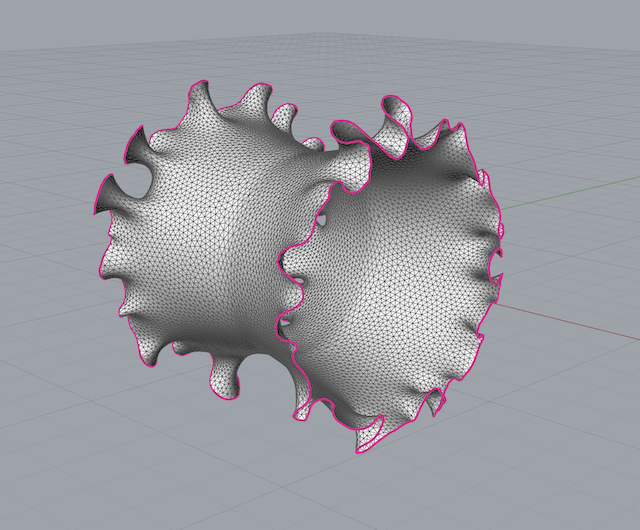} &
        \includegraphics[width=0.15\textwidth]{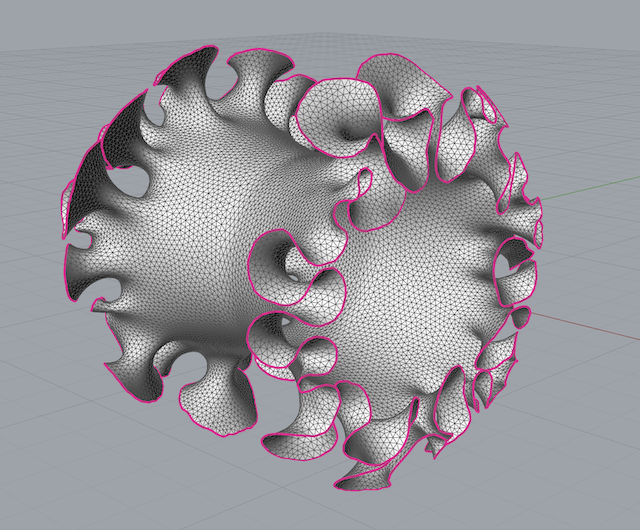} &
        \includegraphics[width=0.15\textwidth]{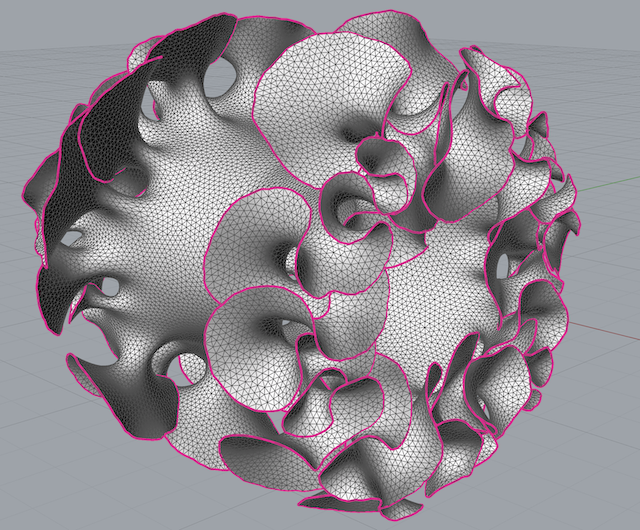} \\
        \includegraphics[width=0.15\textwidth]{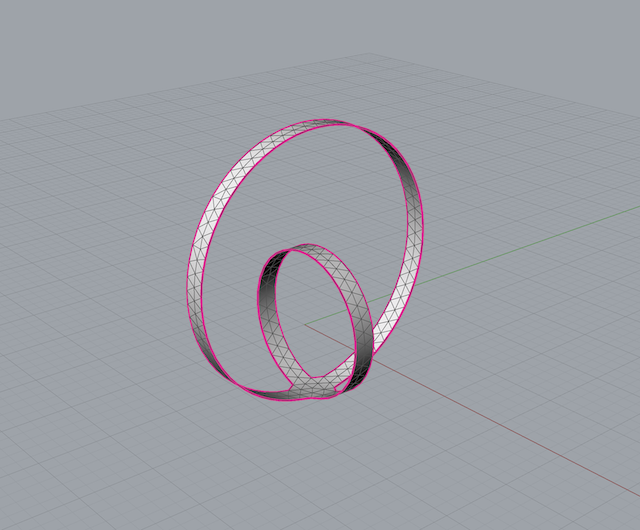} &
        \includegraphics[width=0.15\textwidth]{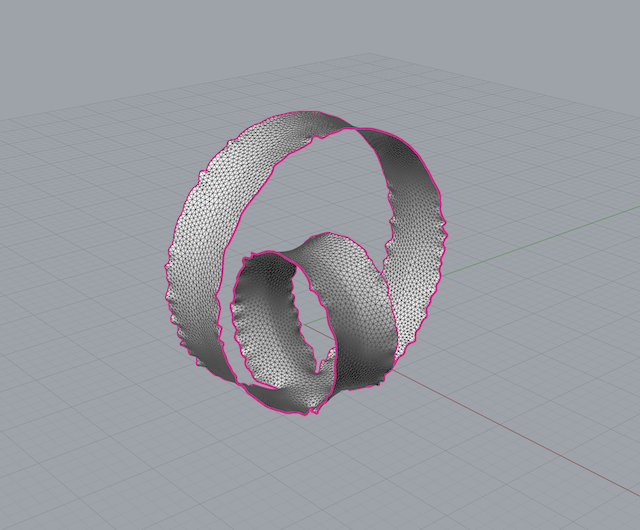} &
        \includegraphics[width=0.15\textwidth]{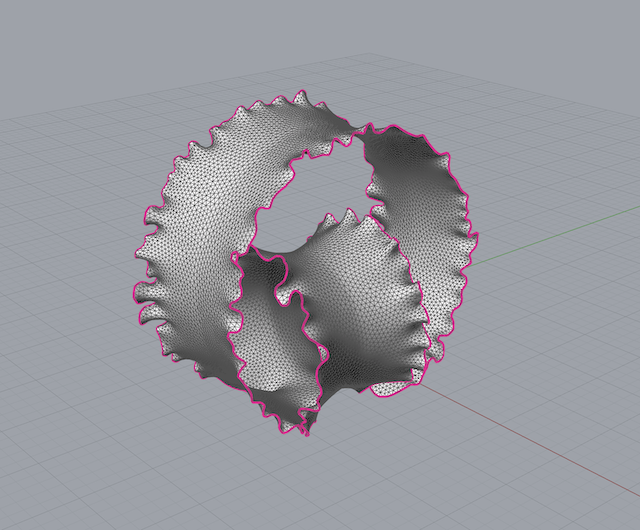} &
        \includegraphics[width=0.15\textwidth]{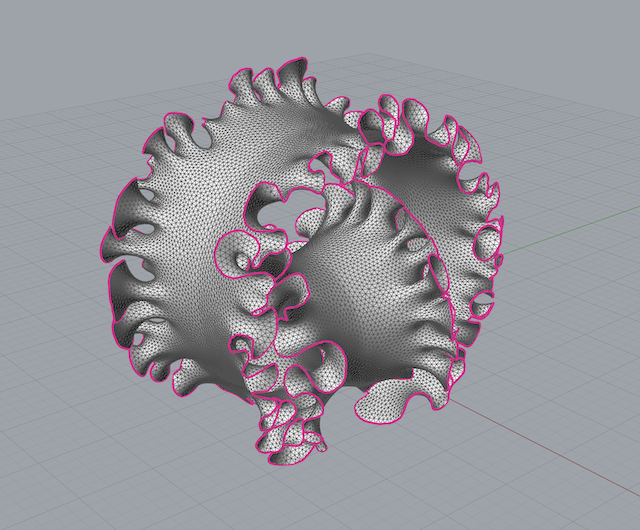} &
        \includegraphics[width=0.15\textwidth]{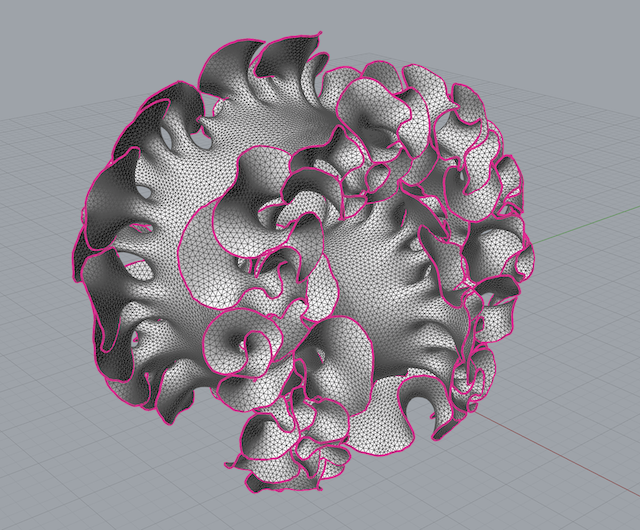} &
        \includegraphics[width=0.15\textwidth]{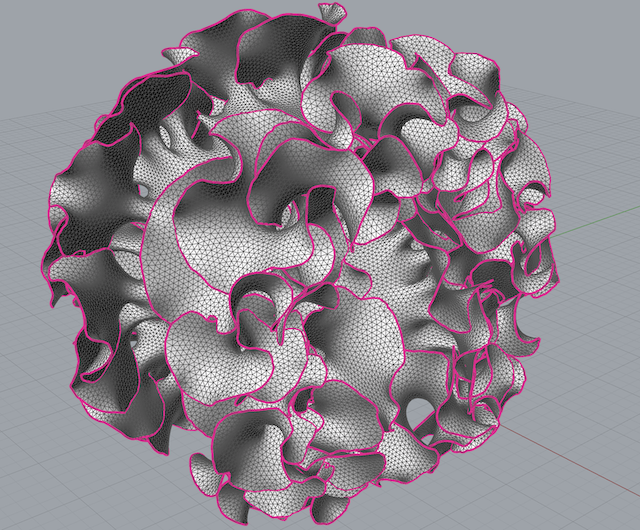} \\
    \end{tabular}
    \caption{Left to right: Morphology development at simulation steps $0, 100, 200, 300, 400, 500$. Top to bottom: initial mesh of momeomorphic types Disk $D^2$, Annulus (Ring, $S^1 \times I$), Möbius Band, and Punctured Torus ($T^2 \backslash D^2$). Boundary edges are highlighted.}
    \label{fig:examples}
\end{figure*}
\begin{table}[h]
    \centering
    \renewcommand{\arraystretch}{1.2} 
    \setlength{\tabcolsep}{1pt} 
    \begin{tabular}{ccccc}
        \includegraphics[width=1.6cm]{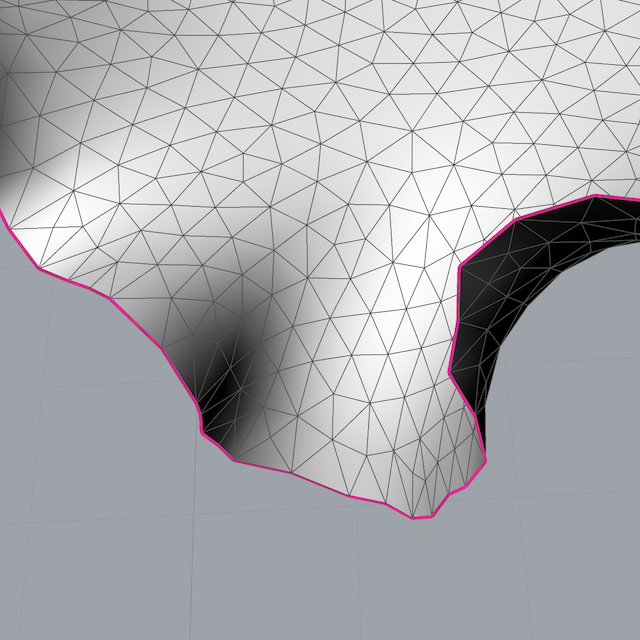} &  
        \includegraphics[width=1.6cm]{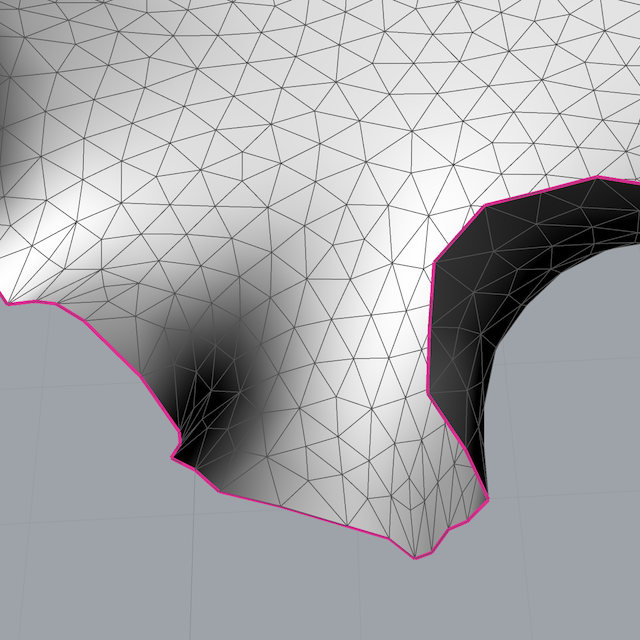} &  
        \includegraphics[width=1.6cm]{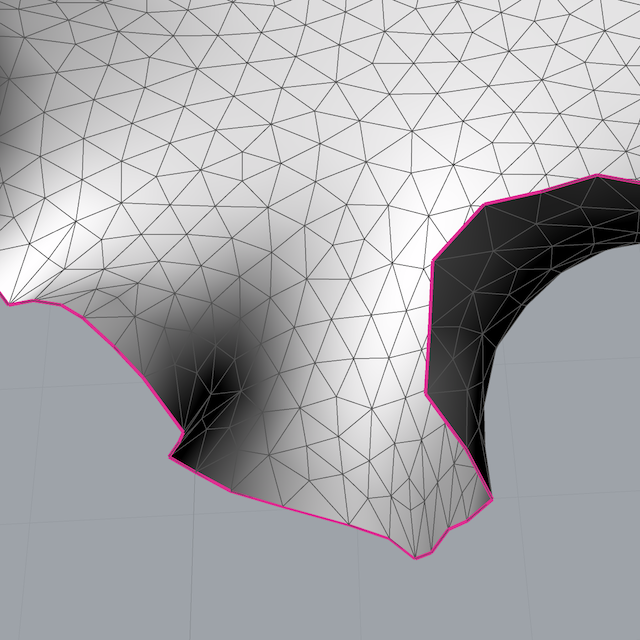} &  
        \includegraphics[width=1.6cm]{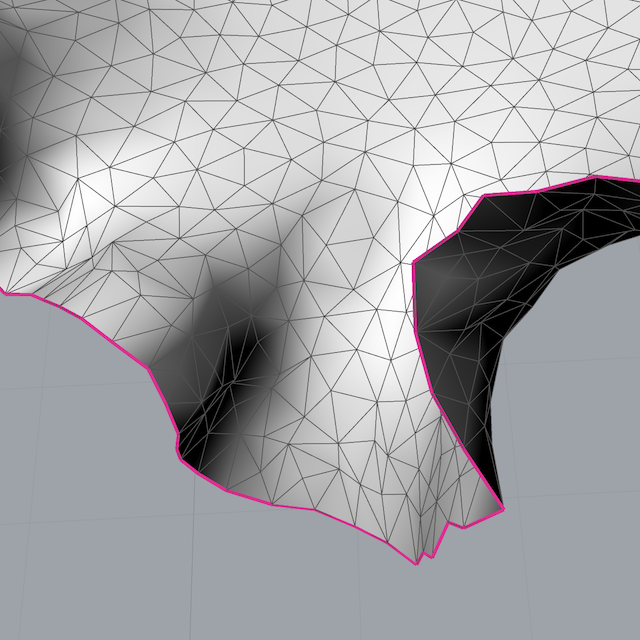} &  
        \includegraphics[width=1.6cm]{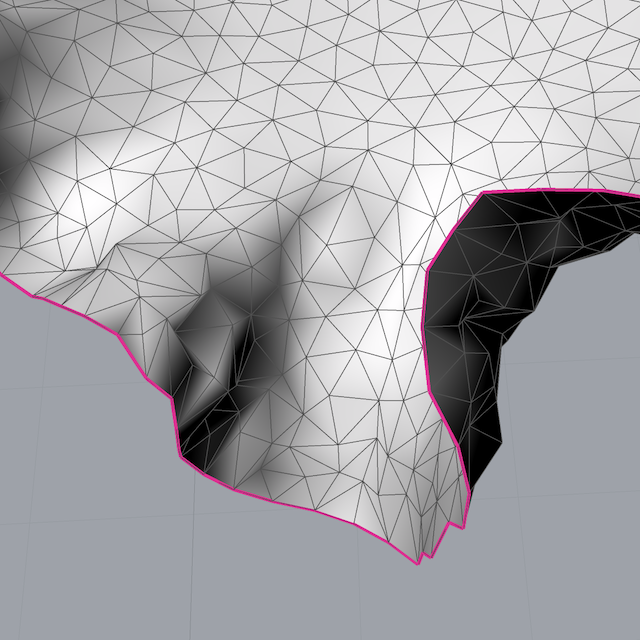} \\  
        init. mesh & (a) & (b) & (c) & (d) 
    \end{tabular}
    \caption{Effect of smoothing and bending forces over $10$ simulation steps. (a) Bending and smoothing forces. (b) Smoothing only. (c) Bending only. (d) Neither.}
    \label{tab:smoothing-and-bending}
\end{table}

\section{Our Method}
\indent \indent Fig~\ref{fig:arch} summarizes our approach. The surface expands as long edges split into shorter edges, creating more faces that are then contracted or enlarged by forces in the simulation. Over time, these local interactions deform the mesh in a concerted manner, creating buckling, branching, and wrapping globally, similar to the curling of leaves and flower petals in nature. These deformations are regularized by smoothing. Meanwhile, corrective collision ensures that the mesh does not self-intersect as the it twists,  turns and folds. Local topology updates improve isotropism and valence.

Cabbage models growth with shell forces. Instead of relying on the Jacobian of shell energy, which is costly to compute and needs a solver, we use the first derivative directly. This allows us to generate a wide range of complex, curling forms within a few minutes in Python in a Jupyter Notebook environment.

Cabbage-Collision is driven by collision forces on each vertex from its local neighborhood. This approximation of shell dynamics generates a related by different family of forms and yield numerous advantages including faster emergence of deformations, with reduced reliance on smoothing and remeshing. The tradeoff is the cost of the growth-driving collision and the corrective collision (which is necessary to prevent self-intersection).

The final triangular result is further processed to preserve its intricate forms with significantly more lightweight geometry and to enable conversion to a closed NURBS volume with crisp, clean edges not present in previous work. This step enables downstream processing in CAD software to build on the simulation result or fabricate it.

\subsection{Shell Growth}
To simulate the tension and stress of the surface, we model edges as compression springs and (for interior edges) torsion springs for faces they lie between. From the current state of the mesh and a desired 'rest' state, we compute shell energy and its first derivative with \cite{BendingForcesAndJacobian, Grinspun:2003DiscreteShells}. 

We set the rest length of the edges to $L_0$, the average length of edges in the initial state of the mesh. This value is held constant for all future simulation steps for consistency. We set the rest area of the mesh faces as that of the equilateral triangle with edge length of $L_0$, as we are aiming for an isotropic mesh. We set the rest angle of each edges (the dihedral angle between the two faces sharing this edge) to be $0$. We note that while this is intuitive for initial meshes that are nearly planar, it is also an effective choice for 3D shapes like rings.

\noindent \textbf{Growth Factors.} We model differential expansion of the 3D surface by splitting edges not only based their current length compared to the ideal $L_0$ but also a user-defined 'growth factor' values that are large in select areas of the mesh. We denote this value as a continuous scalar function $g(v) \in [0, 1]$ of the geodesic distance $d$ from vertex $v$ to the closest 'growth source,' user-defined vertices on the mesh \cite{FloraForm}. $g(v)$ smoothly decays from its nearest growth source \cite{EmYu}. $d$ is normalized to $\tilde{d} \in [0, 1]$ and efficiently approximated with \cite{HeatMethod}.  

\begin{equation}
g(v) =
\begin{cases} 
\left( \frac{\tilde{d}_v}{p} \right)^c p, & \text{if } \tilde{d}_v \leq p, \\
1 - \left( \frac{1 - \tilde{d}_v}{1 - p} \right)^c (1 - p), & \text{if } \tilde{d}_v > p,
\end{cases}
\quad c = \frac{2}{1 - s} - 1
\end{equation}\label{eq:growth-factor}

The cutoff parameter $c$ controls the span of the amplification of normalized geodesic distances closer to $1$. Meanwhile, the steepness parameter $s$ controls the decay of $g(v)$ towards $0$ as the distances decrease.

\noindent \textbf{Growth Sources $S$.} Boundary vertices $\mathcal{V}_b \subset \mathcal{V}$ has the largest impact on the development of the mesh's morphology. During each step, these vertices can be re-assigned, and our method is robust to such changes. That is, the re-assignment of growth sources need not be gradual or smooth. For example, we can assign all boundary vertices of the mesh (which for an open surface there will be more and more of as the mesh expands).

\noindent \textbf{Edge Subdivision.} We split edges based on both its current length and the growth factors of its end point vertices. Splitting long edges serves the dual purpose of ensuring mesh quality (e.g., preventing acute or obtuse triangles) but more importantly expanding the mesh so that there is more of it to be deformed. Additionally, this creates an adaptive refinement effect where at high-curvature areas (e.g. where desired buckling emerges).

\begin{equation}
    \text{Split if} \; \text{length}(e) > \frac{k \ell}{1 + \frac{g(v_1) + g(v_2)}{2}},
    \label{eq:edge-subdivision-criteria}
\end{equation}
where $\ell$ is the current length of the edge between vertices $v_1, v_2$. and $k > 0$ a scalar factor. We use $k = 1$ in our experiments. A larger $k$ does not disrupt the simulation but significantly delays the formation of buckling.

We split boundary edges at its midpoint to create two new shorter edges. An interior edges are subdivided at the weighted average of its end points $v1, v2$ and two opposite vertices $v3, v4$ with Eq.~\eqref{eq:interior_split}. This new vertex does not necessarily lie the edge itself and enhances the smoothness of the mesh \cite{LoopSubdivision}. We note that while splitting interior edges at the midpoint (which is marginally cheaper computation-wise) does not disrupt our method, its introduction of coplanar triangles leads to significantly different morphology. Meanwhile, weighting the position of the newly-created vertex with the growth factor of the edge's end points has little impact and morphology and can introduce very short edges that need to be collapsed to maintain simulation stability.
\begin{equation}
    v_\text{new} = \frac{3}{8}(v_1 + v_2) + \frac{1}{8}(v_3 + v_4).
    \label{eq:interior_split}
\end{equation}
We detect these edges by their ordering in the half-edge data structure of the mesh. We observe that reordering based a `priority' such as their length (or the length amplified by the growth factor of the end points in \cite{eq:edge-subdivision-criteria}) does not enhance growth and can introduce high-valence vertices.

\noindent \textbf{Stretch Forces.}\label{stretch-forces} We compute a spring-like force with Hooke's law for an edge and evenly distribute this force to the edge's two end points.
\begin{equation}
    F_{ij} = (l_{ij} - L_0) * \text{normalize}(v_j - v_i) * \text{stiffness}.
\end{equation}

This force penalizes deviation from the rest length $L_0$ in both the positive and negative directions. In our experiments, we set stiffness to $2$.

\noindent \textbf{Bending Forces.}\label{bending-forces} We compute bending forces with Discrete Shells formulation of bending energy and the derivation in \cite{BendingForcesAndJacobian}. The strength of this force is modulated by a material coefficient that ranges from $5\text{e}-3$ to $3\text{e}-2$ in our experiments. A large bending $k$ delays buckling; a small $k$ expedite that but can lead to visually crumpled results as the dihedral angle of edges increase. 

\noindent\textbf{External Forces.} Additional forces can be introduced to shape growth. We apply gravity to vertices on the mesh \cite{EmYu} as well as a rotational force around a user-defined axis. We scale these forces by their growth factors for a differential effect: areas with low growth factors are pulled down by gravity, while areas with high growth factors gradually orbit around an axis. 

Our smoothing and remeshing play a critical role in the application of forces so that faces of the mesh remains relatively equilateral and do not distort due uneven stretching from the pulling by gravity or twisting by torque.

\noindent\textbf{Remeshing.} We flip edges by the Delaunay criteria so that either opposite vertices to an interior edge lies outside the circumcircle of its opposite triangle. This maximizes the minimum angle of the mesh, preventing the formation of highly obtuse triangles. In practice, we check whether the sum of the sector angle at the two opposite angles is greater than $\pi$ radians, which is a computationally cheaper but just as effective alternative to checking all six angles of the two faces sharing this edge as well as the six new angles of the two faces formed by hypothetically flipping the edge. We flip qualifying edge once per simulation step.

Meanwhile, we collapse edges shorter than $kL_0$ to preserve numerical stability (very stretch or bending forces) and deletes degenerate acute triangles.

Furthermore, we delete 'ears' of the mesh, which are boundary faces made of all boundary vertices. We find that due to such low valence, these faces often become highly acute (hence degenerate) and pulls in nearby vertices to form high-curvature areas. To ensure stability (and additionally for aesthetic realism), we prune these ears by deleting the vertex who has only two neighbors and both of which are boundary, as the interior edge opposite to such vertices cannot be collapsed without creating a loose edge (hence invalid geometry).

We note that these operations sparing adjust the positions and connectivity the geometry of the mesh surface. While simply remeshing the surface completely \cite{Horikawa} can bring far greater improvement to mesh quality, it disrupts the simulation by significantly changing the discretization of the surface (causing a jittering effect over time), removing local anistropism and uneveness that allows differential behavior such as buckling to form over multiple simulation steps.

\subsection{Smoothing} After the application of the sum of forces for each vertex (scaled by a small time step amount of $0.01$ in our experiments), we retain the smoothness and triangle quality of the mesh by feature-away updates to vertex positions.

We move boundary vertices toward the average of their two boundary neighbors. Due to the shrinking nature of this update and the role of boundary areas as a source of buckling behavior, a very small coefficient is used to keep the boundary smooth without smothering emerging features.

We move interior vertices toward a weighted sum of the circumcenter of faces in its one-ring neighborhood. For boundary faces, we use the barycenter as the circumcenter may not lie within the face. The vertex $v$ is thus updated: 

\begin{equation}
  S(v) = \alpha \sum_{f \in \mathcal{F}(v)} \frac{\phi(f)\,\bigl(R(f)^2+t\bigr)}{A(f)},
  \label{eq:smoothing}
\end{equation}
\begin{equation}
  \phi(f) = \chi_{\partial \mathcal{M}}(f)\, c(f) + \bigl(1-\chi_{\partial \mathcal{M}}(f)\bigr)\, b(f),
  \label{eq:phi}
\end{equation}
and the indicator function \( \chi_{\partial \mathcal{M}}(f) \) is given by
\begin{equation}
  \chi_{\partial \mathcal{M}}(f) =
  \begin{cases}
    1, & \text{if } f \in \partial \mathcal{M}, \\
    0, & \text{otherwise.}
  \end{cases}
  \label{eq:indicator}
\end{equation}
where $\alpha \in[0, 1]$ controls the strength of this smoothing (we use $\alpha=0.75$ and $t$ is a small tolerance value to prevent division by zero. \( \mathcal{F}(v) \) is the set of faces incident to vertex \( v \). \( c(f), b(f), R(f), A(f) \) are the circumcenter, barycenter, circumradius, and area of the circumcenter of a face \( f \). \( \partial \mathcal{M} \) denotes the the boundary of the mesh.

\subsection{Preventing Self-Intersection}\label{sec:self-intersection} A non-self-intersecting result is key to both realism as well as downstream geometry processing of the mesh. Advanced methods such as continuous collision detection or energy-based techniques such as \cite{RepulsiveSurfaces}, which require gradient-based optimizations, are accurate and robust but computationally intensive. Instead, we propose an effective alternative with ellipsoidal colliders \cite{FloraForm} for each vertex oriented at its normal. The normal radius $r_n$ is a user-defined value (we used $0.25L_0$) to adjust the separation between the mesh surface and itself, while the tangent radius $r_t$ is the median of all edges sharing this vertex scaled by a factor (we used $0.9$). The choice of median ensures that outliers (e.g. extremely short or long edges) does not skew the size of the ellipsoid.

But unlike previous efforts, we only allow collision to occur between non-adjacent vertices (that do not have an edge between them). This cuts down on computation, but more importantly does not hijack feature formation (see Sec~\ref{sec:collision-driven-growth}). We make this simplification on the observation that two adjacent vertices that are too close to each other (or even overlapping) would simply lead to the edge between them collapsed rather than a self-intersection event. Collision forces from adjacent vertices are redundant given the stretch forces. This simplification is also key to the accuracy of our collision: since the average valence of our mesh is around $6$, their combined collision response can easily overpower that of a non-adjacent vertex. When a vertex $v_i$ start approach such a $v_j$ from afar, there would only be one collision response before $v_i$ gets close enough to also collider with $v_j$'s nearby vertices. Accepting collision events from vertex neighbors means that when $v_i$ is on the brink of being too close to $v_j$, the collision response from $v_j$ simply becomes overpowered (especially if $v_i$ is boundary), leading to a sudden increase in collision response as $v_i$ manages to get very close to $v_j$. For a collision event between $v_i, v_j$, we approximate their penetration of each other by:
\begin{equation}
    \text{collision if} \; p = \delta_n^2 + \delta_t^2 < 1.
    \label{eq:collision-criteria}
\end{equation}
where $\delta_n = \frac{(\mathbf{d}_{ij} \cdot \mathbf{n}_i) \mathbf{n}_i}{\max\{ r_n[i], r_n[j] \}}$ and $\delta_t = \frac{\mathbf{d}_{ij} - \mathbf{d}_{n,i}}{\max\{ r_t[i], r_t[j] \}}$ in which $d_{ij} = v_j - v_i$. The collision force on $v_i$ is then:
\begin{equation}
    F_{ij} = -(k \cdot (1 - p)) \cdot \tilde{d_{ij}},
    \label{eq:collision-force}
\end{equation}
where $\tilde{d_{ij}}$ is the normalized $d_{ij}$ and $k$ is a tunable stiffness parameter. As collision forces are computed and applied every iteration, appropriate choices of $k$ can span $[0.1, 1]$. Since collision response is detected and applied based on penetration, large $k$ can lead to degenerate geometry if surrounding vertices whose ellipsoid did not penetrate any other does not move away, leading to vertices that are very close to one another or even self-intersections. 

Due to the collision-based nature of this approach, it is possible for one vertex to move significantly compared to its neighbors. Therefore, we first smooth the collision force on a vertex toward the average of collision forces on its 1-ring neighbors. After applying this smoothed collision force, we track vertices involved in collision events then smooth neighboring vertices that were not. This ensures that vertices moved apart due to collision were not moved again (hence possibly undone) by the smoothing process. For small $k$, these additional processing were not needed.

\subsection{Enhancing Collision-Driven Growth}\label{sec:collision-driven-growth} Vertex-vertex collision amplify out-of-plane behavior \cite{EmYu, Bachman, Inconvergent, Okunskiy}, creating an effect approximate to stretch and bending forces. Suppose we have uniformly-sized colliders for all vertices on the mesh. For an edge $l_{ij}$ between $v_i, v_j$, collision events between the colliders at these end points ensure that they remain above a certain length. Meanwhile, collision from other vertices keeps $l_{ij}$ from growing too long. While similar to stretch force in principle, the discrete nature of collision response compared to our spring-like formulation in Sec~\ref{stretch-forces}, which is always in effect as long as $l_{ij}$ does not meet the target length $L_0$. Likewise to bending forces in Sec~\ref{bending-forces}, faces on the mesh are pushed toward a configuration where they are nearly equilateral. However, collision alone does not account for the dihedral angle of $l_{ij}$. Despite these limitations, we strengthen this approximation with elements of Cabbage into a viable alternative.

Collision's effectiveness to induce buckling is particularly effective when the collision radius is large enough to envelop vertices beyond the 1-ring. We use spherical colliders with radius $L_0$, which does exactly this for areas with high growth factors. Collision forces are given by:
\begin{equation}
    F_{ij} = k(||d_{ij}|| - L_0)\cdot \text{normalize}(d_{ij}), 
\end{equation}
where $d_{ij} = v_i - v_j$. $k$ is a stiffness factor and we used $k=2$. A further optimization is to only compute collision forces for vertices with high-enough growth factors \cite{EmYu} with the assumption that updates to vertex positions are minimal where growth factors are small. But this threshold needs to be carefully tuned to as low as $0.1$ to control the curvature at very narrow areas of low growth factors. 

\subsection{NURBS Conversion}\label{sec:CAD-conversion} Our approach further distinguishes itself from previous work with a post-processing framework to convert our high-quality mesh into a resolution-invariant CAD format, which is crucial for precise and parametric editing from adding attachment points for screws to geometric patterns, while preserving its details.

We began by quad-remeshing the simulation result $M$ to obtain $M_q$. $M_q$ is then offset by a user-defined amount in either (or both) directions. To supervise this remeshing process, we constrain boundary vertices of $M_q$ to the boundary edges of $M$ and for results with multiscale buckling first smoothly subsidivide $M$ with Loop's method \cite{LoopSubdivision}.

$M_q$'s number of vertices is $20\%$ to $50\%$ that of $M$. With the subdivision algorithm, it could be refined to a very high resolution mesh to ensure smoothness at large scale (e.g. a sculpture for exhibit), removing the faceted appearance of $M$ when viewed extremely up-close due to its discrete nature. Furthermore, $M_q$ can be converted into NURBS format \cite{Rhino}, which is naturally suited to operations such as boolean at high precision (e.g. drilling a hole with a radius, chamfers, and clearance). This conversion ensures that our simulation results are not just visually appealing renderings in the virtual world but concrete products that provides a strong foundation for further design and manufacturing.

\subsection{Implementation Detail}
We used the OpenMesh \cite{OpenMesh} package's python binding for its half-edge data structure. For \cite{HeatMethod}, we used the potpourri3d package \cite{potpourri3d}. The simulation process can be fully run on in a Jupyter notebook environment on a personal laptop (we used an Apple laptop with 16GB RAM). We conducted the conversion to NURBS is carried out in Grasshopper \cite{Rhino} but note that open-source projects like Blender \cite{Blender} readily provide very similar functionality. 

\begin{table}[h]
    \centering
    \resizebox{\columnwidth}{!}{
    \begin{tabular}{lcccc}
        \toprule
        \textbf{Method} & \textbf{Task Time (s) $\downarrow$} & \textbf{SUS Score $\uparrow$} & \textbf{NASA-TLX $\uparrow$} & \textbf{Aesthetic Rating $\uparrow$} \\
        \midrule
        \multicolumn{5}{c}{\textbf{All Participants}} \\
        \textbf{Cabbage (Ours)}  & \textbf{45.2 $\pm$ 5.1}  & \textbf{88.4 $\pm$ 4.2}  & \textbf{22.3 $\pm$ 3.5}  & \textbf{6.7 $\pm$ 0.3} \\
        Yu \cite{EmYu}       & 60.7 $\pm$ 8.3  & 75.2 $\pm$ 5.1  & 35.4 $\pm$ 4.2  & 5.8 $\pm$ 0.5 \\
        Okunskiy \cite{Okunskiy} & 70.5 $\pm$ 10.2 & 70.8 $\pm$ 6.4  & 40.1 $\pm$ 5.3  & 5.3 $\pm$ 0.6 \\
        Horikawa \cite{Horikawa} & 90.9 $\pm$ 12.1 & 65.6 $\pm$ 7.3  & 50.5 $\pm$ 6.2  & 4.8 $\pm$ 0.7 \\
        Bachman \cite{Bachman}  & 80.4 $\pm$ 10.3 & 68.3 $\pm$ 6.1  & 45.7 $\pm$ 5.4  & 5.0 $\pm$ 0.5 \\
        \midrule
        \multicolumn{5}{c}{\textbf{Technical Users}} \\
        \textbf{Cabbage (Ours)}  & \textbf{40.3 $\pm$ 4.2}  & \textbf{90.1 $\pm$ 3.3}  & \textbf{20.5 $\pm$ 2.4}  & \textbf{6.8 $\pm$ 0.2} \\
        Yu \cite{EmYu}       & 55.8 $\pm$ 7.1  & 78.6 $\pm$ 4.2  & 33.2 $\pm$ 3.5  & 5.9 $\pm$ 0.4 \\
        Okunskiy \cite{Okunskiy} & 65.4 $\pm$ 9.3  & 72.5 $\pm$ 5.4  & 38.6 $\pm$ 4.2  & 5.4 $\pm$ 5.5 \\
        Horikawa \cite{Horikawa} & 85.2 $\pm$ 10.6 & 68.9 $\pm$ 6.3  & 48.7 $\pm$ 5.1  & 4.9 $\pm$ 0.6 \\
        Bachman \cite{Bachman}  & 75.6 $\pm$ 9.2  & 70.4 $\pm$ 5.2  & 43.1 $\pm$ 4.3  & 5.1 $\pm$ 0.4 \\
        \midrule
        \multicolumn{5}{c}{\textbf{Non-Technical Users}} \\
        \textbf{Cabbage (Ours)}  & \textbf{50.6 $\pm$ 6.1}  & \textbf{86.2 $\pm$ 5.3}  & \textbf{24.8 $\pm$ 3.6}  & \textbf{6.6 $\pm$ 0.4} \\
        Yu \cite{EmYu}       & 65.3 $\pm$ 8.4  & 73.5 $\pm$ 5.1  & 37.1 $\pm$ 4.3  & 5.7 $\pm$ 0.5 \\
        Okunskiy \cite{Okunskiy} & 75.7 $\pm$ 10.5 & 68.9 $\pm$ 6.2  & 42.3 $\pm$ 5.4  & 5.2 $\pm$ 0.6 \\
        Horikawa \cite{Horikawa} & 95.2 $\pm$ 12.3 & 63.8 $\pm$ 7.2  & 52.6 $\pm$ 6.3  & 4.7 $\pm$ 0.7 \\
        Bachman \cite{Bachman}  & 85.9 $\pm$ 10.7 & 66.4 $\pm$ 6.1  & 47.5 $\pm$ 5.2  & 4.9 $\pm$ 0.5 \\
        \bottomrule
    \end{tabular}}
    \caption{User study results for all participants, technical users (CS students and industry professionals with coding experiences), and non-technical users (designers and artists). Lower task time and NASA-TLX indicate better performance, while higher SUS and aesthetic scores are preferable.}
    \label{tab:user-study}
\end{table}

\begin{table}[h]
    \centering
    \renewcommand{\arraystretch}{1.2} 
    \setlength{\tabcolsep}{2pt} 
    \begin{tabular}{lccccc}
        \toprule
        \textbf{Method} & (1) $\downarrow$ & (2) $\downarrow$ & (3) $\uparrow$ & (4) $\downarrow$ & (5) $\downarrow$ \\
        \midrule
        Yu \cite{EmYu} & 5  & \textbf{0}  & 0.89  & 0.32  & 6.0  \\
        Okunskiy \cite{Okunskiy} & 20  & 15  & 0.65  & 0.9  & 6.9  \\
        Horikawa \cite{Horikawa} & 2 & \textbf{0} & \textbf{0.95}  & 0.25  &  6.0 \\
        Bachman \cite{Bachman} & 15  & 5 & 0.985  & 0.8  & 6.4  \\
        \textbf{Cabbage (Ours)} & \textbf{0}  & \textbf{0}  & \textbf{0.95}  & \textbf{0.2}  & \textbf{5.9}  \\
        \bottomrule
    \end{tabular}
    \caption{Comparison of five metrics: (1) percentage of self-intersection events, (2) percentage of simulation failure, (3) average face quality of final result, (4) squared sum of dihedral angle of final result (radians), and (5) average valence of final result.}
    \label{tab:comparison}
\end{table}

\section{Experiments}
\subsection{Dataset and Evaluation Metric} We compare Cabbage to Yu \cite{EmYu}, Okunskiy \cite{Okunskiy}, Horikawa \cite{Horikawa}, and Bachman \cite{Bachman} which are the only methods with reproducible implementations to the best of our knowledge. We evaluate these methods on dataset of $120$ randomly-generated meshes equally distributed over three homeomorphic types of 3D open surfaces: Disk $D^2$, Annulus (Ring, $S^1 \times I$), a Möbius Band, and a Punctured Torus ($T^2 \backslash D^2$). Due to the distinct properties of the methods we compare, the resulting mesh from running a constant number of iterations can vary greatly. Therefore, we instead set a goal of $3,000$ vertices beyond which we stop the simulation. For the experiments, we record the occurrence of self-intersection events, growth failure, and distince morphological development. We also conduct a user study in Tab~\ref{tab:user-study} to meter the real-world effectiveness of Cabbage. 

\subsection{Quantitative Results} Tab~\ref{tab:comparison} reports mesh quality metrics during simulation.

\noindent \textbf{Self-Intersection}. In Yu and Bachman, collision responses serve to both induce buckling and prevent self-intersections, which is less effective in crowded areas due to jagged boundaries and large dihedral angles. Okunskiy’s spherical colliders can leave gaps in degenerate triangles, causing intersections. Cabbage’s dedicated corrective collision—applied after vertex adjustments—combined with adaptive ellipsoidal colliders and additional smoothing, effectively prevents self-intersections.

\noindent \textbf{Simulation Failure}. In collision-driven approaches, extremely large collision responses, especially after self-intersection events, where the part of the mesh that penetrated another part of itself is pushed in a direction opposite to the part that is not, distorts the mesh, leading to degenerate faces with very long edges.  

\noindent \textbf{Face Quality}. Okunskiy’s method yields sharp triangles due to a lack of topology updates, hindering collision effectiveness in high-curvature regions. Horikawa’s isotropic remeshing and Bachman’s edge constraints preserve face shape, while Yu’s use of collision, edge flips, and barycentric smoothing struggles when low-growth areas buckle. In contrast, Cabbage employs stretch forces to drive edge lengths toward $L_0$ and uses ODT smoothing for isotropism; Cabbage-Collision decouples smoothing from growth factor distribution, ensuring consistent face quality.

\noindent \textbf{Vertex Valence}. The absence of edge flips in Okunskiy and Bachman results in high vertex valence, whereas Horikawa’s global remeshing produces more uniform valence. Yu and Cabbage limit valence through edge flips, with Cabbage further reducing it via edge collapses.

\noindent \textbf{Mesh Smoothness}. \label{sec:smoothness} Methods differ notably in smoothness. Okunskiy’s meshes are jagged from the start; Bachman’s smoothness degrades as the mesh grows rapidly. Horikawa’s global remeshing and Laplacian smoothing, though conceptually effective, yield crumpled, jittery surfaces. Yu’s barycentric smoothing preserves smoothness but can suppress local buckling details when overused. Cabbage’s use of distinct bending forces alongside feature-preserving ODT smoothing achieves and maintains a consistently smooth mesh conducive to controlled buckling.

\subsection{Qualitative Results}. 
We conducted a user study with $83$ domain experts and students in fine art, 3D art, and digital fabrication to compare Cabbage with existing methods in aesthetic quality, design potential, and manufacturability. Our evaluation metrics are the task completion time, SUS score $(0\text{–}100)$, NASA-TLX $(0\text{–}100)$, and an aesthetic rating $(0\text{–}7)$. Tab~\ref{tab:user-study} highlights Cabbage's lead across these metrics.

\noindent \textbf{Aesthetics and Morphological Range}. Cabbage generates high-quality meshes with a diverse range of shapes. Its control over growth factor distribution enables differential growth not available in Bachman’s method—which converges into bush-like forms—or in Horikawa’s approach that produces large, wrinkly meshes through isotropic remeshing. While Okunskiy’s Blender plugin offers user-defined vertex weights, its mesh quality limits realism, and Yu’s method, though smooth, lacks nuanced control over buckling. In contrast, Cabbage preserves mesh smoothness while allowing intuitive adjustment via bending stiffness.

\noindent \textbf{Design and Fabrication}. Cabbage prevents self-intersections on large meshes (up to $1 \times 10^5$ vertices) and incorporates a CAD conversion pipeline that produces lightweight, editable formats. This feature is especially effective for smooth, high-resolution meshes, overcoming the challenges of distinguishing sharp features from curved forms in coarse-grain buckling regions.

\subsection{Ablation} \noindent \textit{Growth Factor Distribution}. The growth cutoff critically influences morphology by confining buckling to regions with high growth factors. With low steepness and sparse factors, nearby regions quickly separate, forming elongated structures while restricting buckling. Steepness has little impact.

\noindent \textit{Bending Force}. Bending forces are essential for maintaining a smooth, feature-preserving surface during buckling (see Tab~\ref{tab:smoothing-and-bending}). However, a constant high bending coefficient delays out-of-plane deformations. By starting with a low coefficient and increasing it over time, buckling emerges sooner, especially for planar initial meshes.

\noindent \textit{Initial Surface}. The final morphology is heavily influenced by the initial mesh’s geometry and topology. Nearly isotropic initial meshes allow a wider range of stable growth parameters, though high-valence vertices from the original state may persist.

\section{Conclusion}
Cabbage is a novel, open-source framework that efficiently models the buckling of 3D open surfaces, as exemplified by the curling of flower petals. Through edge subdivision, shell force-driven expansion, feature-aware remeshing, and robust corrective collision, Cabbage outperforms state-of-the-art methods by generating high-quality, self-intersection-free triangular meshes and CAD-ready surfaces, enriching digital design and  fabrication.
{
    \small
    \bibliographystyle{ieeenat_fullname}
    \bibliography{main}
}

\end{document}